\documentclass[11pt,a4paper]{article}
\usepackage[hyperref]{conll2018}
\usepackage{times}
\usepackage{latexsym}

\usepackage{url}
\usepackage{graphicx}
\usepackage{gb4e}
\noautomath
\exewidth{(\thexnumi)}
\usepackage{amsmath}
\usepackage{amssymb}
\usepackage{qtree}
\usepackage{booktabs}

\aclfinalcopy 

\newcommand{\citepos}[1]{\citeauthor{#1}'s (\citeyear{#1})}

\title{The Lifted Matrix-Space Model for Semantic Composition}

\author{
	WooJin Chung$^{1}$\\
	{\tt woojin@nyu.edu}
	\And
	Sheng-Fu Wang$^{1}$\\
	{\tt shengfu.wang@nyu.edu}
	\And
	Samuel R.~Bowman$^{1,2}$\\
	{\tt bowman@nyu.edu}
	\AND
	$^{1}$\normalfont Dept. of Linguistics\\New York University\\10 Washington Place\\New York, NY 10003\And
	$^{2}$\normalfont Center for Data Science\\New York University\\60 Fifth Avenue\\New York, NY 10011
}

\date{}

\begin{document}
\maketitle
\begin{abstract}
Tree-structured neural network architectures for sentence encoding draw inspiration from the approach to semantic composition generally seen in formal linguistics, and have shown empirical improvements over comparable sequence models by doing so.
Moreover, adding multiplicative interaction terms to the  composition functions in these models can yield significant further improvements. However, existing compositional approaches that adopt such a powerful composition function scale poorly, with parameter counts exploding as model dimension or vocabulary size grows. We introduce the Lifted Matrix-Space model, which uses a global transformation to map vector word embeddings to matrices, which can then be composed via an operation based on matrix-matrix multiplication. Its composition function effectively transmits a larger number of activations across layers with relatively few model parameters. We evaluate our model on the Stanford NLI corpus, the Multi-Genre NLI corpus, and the Stanford Sentiment Treebank and find that it consistently outperforms TreeLSTM \cite{Tai2015}, the previous best known composition function for tree-structured models.
\end{abstract}

\section{Introduction}
\label{sec:introduction}
Contemporary theoretical accounts of natural language syntax and semantics consistently hold that sentences are tree-structured, and that the meaning of each node in each tree is calculated from the meaning of its child nodes using a relatively simple \textit{semantic composition} process which is applied recursively bottom-up \citep{Chierchia1990, Dowty2007}. In tree-structured recursive neural networks \citep[TreeRNN;][]{Socher2010}, a similar procedure is used to build representations for sentences for use in natural language understanding tasks, with distributed representations for words repeatedly fed through a neural network \textit{composition function} according to a binary tree structure supplied by a parser. The success of a tree-structured model largely depends on the design of its composition function. 

\begin{figure}[t]
	\begin{center}
		\includegraphics[width=7.6cm]{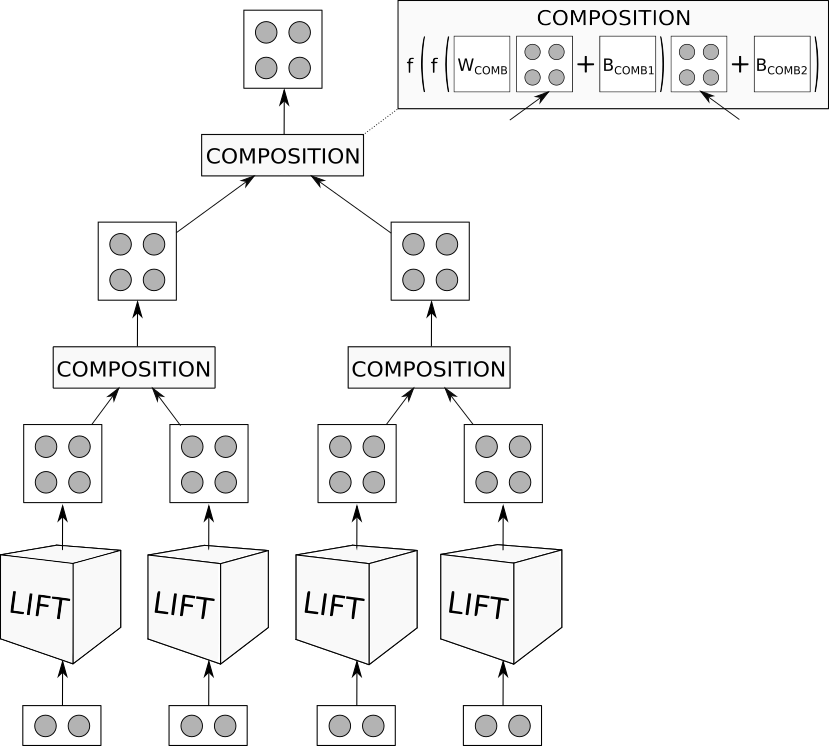}
		\caption{The Lifted Matrix-Space model in schematic form. Words are stored as vectors and projected into matrix space by the \textsc{lift} layer. A parametric \textsc{composition} function combines pairs of these matrices using multiplicative interactions.}
		\label{fig:lms}
	\end{center}
\end{figure}

It has been repeatedly shown that a composition function that captures multiplicative interactions between the two items being composed yields better results \cite{Rudolph2010, Socher2012, Socher2013} than do otherwise-equivalent functions based on simple linear interactions. This paper presents a novel model which advances this line of research, the Lifted Matrix-Space model. We utilize a tensor-parameterized \textsc{lift} layer that learns to produce matrix representations of words that are dependent on the content of pre-trained word embedding vectors. Composition of two matrix representations is carried out by a composition layer, into which the two matrices are sequentially fed. Figure \ref{fig:lms} illustrates the model design.

Our model was inspired by Continuation Semantics \cite{Barker2014, Charlow2014}, where each symbolic representation of words is converted to a higher-order function. There is a consensus in linguistic semantics that a subset of natural language expressions correspond to higher-order functions. Inspired by the works in programming language theory, Continuation Semantics takes a step further and claims that all expressions have to be converted into a higher-order function before they participate in semantic composition. The theory bridges a gap between linguistic semantics and programming language theory, and reinterprets various linguistic phenomena from the view of computation. While we do not directly implement Continuation Semantics, we follow its rough contours: We convert low-level representations (vectors) to higher-order functions (matrices), and composition only takes place between the higher-order functions.

A number of models have been developed to capture the multiplicative interactions between distributed representations. While having a similar objective, the proposed model requires fewer model parameters than the predecessors because it does not necessarily learn each word matrix representation separately, and the number of parameters for the composition function is not proportional to the cube of the hidden state dimension. Because of this, it can be trained with larger vocabularies and more hidden state activations than was possible with its predecessors.

We evaluate our model primarily on the task of \textit{natural language inference} \citep[NLI;][]{MacCartney2009}. 
The task consists in determining the inferential relation between a given pair of sentences. It is a principled and widely-used evaluation task for natural language understanding, and knowing the inferential relations is closely related to understanding the meaning of an expression \cite{Chierchia1990}. While other tasks such as question answering or machine translation require a model to learn task-specific behavior that goes beyond understanding sentence meaning, NLI results highlight sentence understanding performance in isolation. We also include an evaluation on sentiment classification for comparison with some earlier work.

We find that our model outperforms existing approaches to tree-structured modeling on all three tasks, though it does not set the state of the art on any of them, falling behind other types of complex model. We nonetheless expect that this method will be a valuable ingredient in future models for sentence understanding and a valuable platform for research of compositionality in learned representations.

\section{Related work} 
\label{sec:related works}

Composition functions for tree-structured models have been thoroughly studied in recent years \cite{Mitchell2010, Baroni2010, Zanzotto2010, Socher2011}. While this line of research has been successful, the majority of the existing models ultimately rely on the additive linear combination of vectors. The Tree-structured recursive neural networks (TreeRNN) of \citet{Socher2010} compose two child node vectors $\vec{h}_{l}$ and $\vec{h}_{r}$ using this method:

\begin{exe}
	\ex
	$\vec{h} = tanh(W \left[ \begin{array}{c} \vec{h}_{l}\\ \vec{h}_{r} \end{array} \right] + \vec{b})$
	\label{treernn2}
\end{exe}
where $\vec{h}_{l}$, $\vec{h}_{r}$, $\vec{h}$, $\vec{b}$ $\in \mathbb{R}^{\text{$d \times 1$}}$, and $W \in \mathbb{R}^{\text{$d \times 2d$}}$. Throughout this paper, $d$ stands for the number of activations of a given model.

However, there is no reason to believe that the additive linear combination of vectors is adequate for modeling semantic composition. Formal work in linguistic semantics has shown that many linguistic expressions are well-represented as functions. Accordingly, composing two meanings typically require feeding an argument into a function \citep[function application;][]{Heim1998}. Such an operation involves a complex interaction between the two meanings, but the classic TreeRNN does not supply any additional means to capture the interaction.

\citet{Rudolph2010} report that matrix multiplication, as opposed to element-wise addition,  is more suitable for semantic composition. Their Compositional Matrix-Space model (CMS) represents words and phrases as matrices, and they are composed via a simple matrix multiplication:

\begin{exe}
	\ex $P = AB$
\end{exe}
where $A$, $B$, $P$ $\in \mathbb{R}^{\text{d $\times$ d}}$ are matrix representations of the word embeddings. They provide a formal proof that element-wise addition/multiplication of vectors can be subsumed under matrix multiplication. Moreover, they claim that the order-sensitivity of matrix multiplication is adequate for capturing the semantic composition because natural language is order-sensitive.

However, as \citet{Socher2012} note, CMS loses syntactic information during composition due to the associative character of matrix multiplication. For instance, the following two tree structures in (\ref{tree}) are syntactically distinct, but CMS would produce the same result for both structures because its mode of composition is associative.

\begin{exe}
	\item a.~\Tree[A [ B C ] ]~~~~~~~b.~\Tree[ [ A B ] C ]\label{tree}
\end{exe}

CMS cannot distinguish the meaning of the two tree structures and invariably produces ABC (a sequence of matrix multiplications). Therefore, the information on syntactic constituency would be lost. This makes the model less desirable for handling semantic composition of natural language expressions for two reasons: First, the principle of compositionality is violated. Much of the success of the tree-structured models can be credited to the shared hypothesis that the meaning of every tree node is derived from the meanings of its child nodes. Abandoning this principle of compositionality gives up the advantage. Second, it cannot handle structural ambiguities exemplified in (\ref{ambiguity}).

\begin{exe}
	\ex John saw a man with binoculars.
	\label{ambiguity}
	\ex
	a.\hspace{-1pt}~\Tree[John [ [ saw  [ a man ] ] [ with binoculars ] ] ]\\
	b.\hspace{-1pt}~\Tree[John [ saw  [ a [ man [ with binoculars ] ] ] ] ]
\end{exe}

The sentence has two interpretations that can be disambiguated with the following paraphrases: (i) John saw a man via binoculars, and (ii) John saw a man who has binoculars. The common syntactic analysis of the ambiguity is that the prepositional phrase \textit{with binoculars} can attach to two different locations. If it attaches to the verb phrase \textit{saw a man}, the first interpretation arises. On the other hand, if it attaches to the noun \textit{man}, the second interpretation is given. However, if the structural information is lost, we would have no way to disambiguate the two readings.

\citepos{Socher2012} Matrix-Vector RNN (MV-RNN) is another attempt to capture the multiplicative interactions between two vectors, while conforming to the principle of compositionality. They hypothesize that representing operators as matrices can better reflect operator semantics. For each lexical item, a matrix (trained parameter) is assigned in addition to the pre-trained word embedding vector. The model aims to assign the right matrix representations to operators while assigning an identity matrix to words with no operator meaning. One step of semantic composition is defined as follows:

\begin{exe}
	\ex
	$\vec{h} = f(B \vec{a}, A \vec{b}) = g(W \left[ \begin{array}{c} B \vec{a}\\ A \vec{b} \end{array} \right])$
	\ex
	$H = f_{M}(A, B) = W_{M} \left[ \begin{array}{c} A\\ B \end{array} \right]$
	\label{mvrnn}
\end{exe}
where $\vec{a}$, $\vec{b}$, $\vec{h}$ $\in \mathbb{R}^{\text{d $\times$ 1}}$, $A$, $B$, $H$ $\in \mathbb{R}^{\text{d $\times$ d}}$, and $W$, $W_{M}$ $\in \mathbb{R}^{d \times 2d}$.

MV-RNN is computationally costly as it needs to learn an additional $d \times d$ matrix for each lexical item. It is empirically known that the size of the vocabulary is roughly proportional to the size of the corpus \citep[Heaps' law;][]{herdan1960type}, therefore the number of model parameters increases as the corpus gets bigger. This makes the model less ideal for handling a large corpus: having a huge number of parameters causes a problem both for memory usage and for learning efficiency.

\citet{Chen2013} and \citet{Socher2013} present the recursive neural tensor network (RNTN) which reduces the computational complexity of MV-RNN, while capturing the multiplicative interactions between child vectors. The model introduces a third-order tensor \textbf{\textit{V}} which interacts with the child node vectors as follows:

\begin{exe}
	\ex
	$\vec{h} = tanh(W \left[ \begin{array}{c} \vec{h}_{l}\\ \vec{h}_{r} \end{array} \right] + \vec{b} + \vec{h}_{l}^{T} \textbf{V} \vec{h}_{r})$
	\label{ntn}
\end{exe}

\noindent where $\vec{h}_{l}$, $\vec{h}_{r}$, $\vec{b}$ $\in \mathbb{R}^{\text{$d \times 1$}}$, W $\in \mathbb{R}^{\text{$d \times 2d$}}$, and \textbf{V} $\in \mathbb{R}^{\text{$d \times d \times d$}}$. RNTN improves on MV-RNN in that its parameter size is not proportional to the size of the corpus. However, the addition of the third-order tensor \textbf{V} of dimension $d \times d \times d$ still requires proportionally more parameters.

\begin{table*}[t]
	\small
	\begin{center}
		\begin{tabular}{ lcccc }
			\toprule
			Model & Params. & Associative & Multiplicative & Activation size w.r.t. TreeRNN\\
			\midrule
			TreeRNN/LSTM & O($d \times d$) & No & No & $1$\\
			CMS & O($V\times d \times d$) & Yes & Yes & $1/V$\\
			MV-RNN & O($V\times d \times d$) & No & Yes & $1/V$\\
			RNTN & O($d \times d \times d$) & No & Yes & $1/d$\\
			\midrule
			\textbf{LMS} (this work) & O($d \times d_{emb}$) & No & Yes & $1/d_{emb}$\\
			\bottomrule
		\end{tabular}
		\caption{Summary of the models. \textit{Params.} is the number of model parameters (not counting pretrained word vectors), $d$ is the number of activations at each tree node, $d_{emb}$ is the dimension of the word embeddings, and V is the size of the vocabulary. \textit{Associative} and \textit{Multiplicative} indicate whether composition is associative and whether it includes multiplicative interactions between inputs, respectively. \textit{Activation size w.r.t. TreeRNN} shows how activation sizes scale with respect to TreeRNN when all of the models have the same parameter count.}
		\label{tab:summary}
	\end{center}
\end{table*}

The last composition function relevant to this paper is the Tree-structured long short-term memory networks \citep[TreeLSTM;][]{Tai2015, Zhu2015, Le2015}, particularly the version over a constituency tree. It is an extension of TreeRNN which adapts long short-term memory \citep[LSTM;][]{Hochreiter1997} networks. It shares the advantage of LSTM networks in that it prevents the vanishing gradient problem \cite{Hochreiter2001}.

Unlike TreeRNN, the output hidden state $\vec{h}$ of TreeLSTM is not directly calculated from the hidden states of its child nodes, $\vec{h}_{l}$ and $\vec{h}_{r}$. Rather, each node in TreeLSTM maintains a cell state $\vec{c}$ that keeps track of important information of its child nodes. The output hidden state $\vec{h}$ is drawn from the cell state $\vec{c}$ by passing it through an output gate $\vec{o}$.

The cell state is calculated in three steps: (i) Compute a new candidate $\vec{g}$ from $\vec{h}_{l}$ and $\vec{h}_{r}$. TreeLSTM selects which values to take from the new candidate $\vec{g}$ by passing it through an input gate $\vec{i}$. (ii) Choose which values to forget from the cell states of the child nodes, $\vec{c}_{l}$ and $\vec{c}_{r}$. For each child node, an element-wise product ($\odot$) between its cell state and the forget gate (either $\vec{f}_{l}$ and $\vec{f}_{r}$, depending on the child node) is calculated. (iii) Lastly, sum up the results from (i) and (ii).

\begin{exe}
	\ex
	$\vec{g} = tanh \left( W \left[ \begin{array}{c} \vec{h}_{l}\\ \vec{h}_{r} \end{array} \right] + \vec{b} \right)$
	\ex
	$\left[ \begin{array}{c} \vec{i}\\ \vec{f}_{l}\\ \vec{f}_{r}\\ \vec{o} \end{array} \right]
	= \left[ \begin{array}{c} \sigma\\ \sigma\\ \sigma\\ \sigma \end{array} \right]
	\left( W \left[ \begin{array}{c} \vec{h}_{l}\\ \vec{h}_{r} \end{array} \right] + \vec{b} \right)$
	\ex
	$\vec{c} = \vec{f}_{l} \odot \vec{c}_{l} + \vec{f}_{r} \odot \vec{c}_{r} + \vec{i} \odot \vec{g}$
	\ex
	$\vec{h} = \vec{o} \odot tanh(\vec{c})$
\end{exe}

TreeLSTM achieves the state-of-the-art performance among the tree-structured models in various tasks, including natural language inference and sentiment classification. However, there are non-tree-structured models on the market that outperform TreeLSTM. Our goal is to design a stronger composition function that enhances the performance of tree-structured models. We develop a composition function that captures the multiplicative interaction between distributed representations. At the same time, we improve on the predecessors in terms of scalability, making the model more suitable for larger datasets.

To recapitulate, TreeRNN and TreeLSTM reflect the principle of compositionality but cannot capture the multiplicative interaction between two expressions. In contrast, CMS incorporates multiplicative interaction but violates the principle of compositionality. MV-RNN is compositional and also captures the multiplicative interaction, but it requires a learned $d \times d$ matrix for each vocabulary item. RNTN is also compositional and incorporates multiplicative interaction, but it requires less parameters than MV-RNN. Nevertheless, it requires significantly more parameters than TreeRNN or TreeLSTM. Table~\ref{tab:summary} is an overview of the discussed models.

Other interesting works enrich semantic composition with additional context such as grammatical roles or function/argumenthood \cite{Clark2008, Erk2008, Grefenstette2014, Asher2016, Weir2016}.

\section{The Lifted Matrix-Space model}
\label{sec:proposal}

\subsection{Base model}

We present the Lifted Matrix-Space model (LMS) which renders semantic composition in a novel way. Our model consists of three subparts: the \textsc{lift} layer, the composition layer, and the TreeLSTM wrapper. The \textsc{lift} layer takes a word embedding vector and outputs a corresponding $\sqrt{d} \times \sqrt{d}$ matrix (eq. \ref{lift layer}).

\begin{exe}
	\ex
	$H = tanh(\textbf{W}_{\textsc{lift}} \vec{c} + B_{\textsc{lift}})$
	\label{lift layer}
\end{exe}
where $\vec{c} \in \mathbb{R}^{d_{emb} \times 1}$, $B_{\textsc{lift}} \in \mathbb{R}^{\sqrt{d} \times \sqrt{d}}$, and $\textbf{W}_{\textsc{lift}} \in \mathbb{R}^{\sqrt{d} \times \sqrt{d} \times d_{emb}}$. The resulting $H$ matrix serves as an input for the composition layer.

Given the matrix representations of two child nodes, $H_{l}$ and $H_{r}$, the composition layer first takes $H_{l}$ and returns a hidden state $H_{inner} \in \mathbb{R}^{\sqrt{d} \times \sqrt{d}}$ (eq. \ref{ex:composition1}). Since $H_{inner}$ is also a matrix, it can function as the weight matrix for $H_{r}$. The composition layer multiplies $H_{inner}$ with $H_{r}$, adds a bias, and feeds the result into a non-linear activation function (eq. \ref{ex:composition2}). This yields $H_{cand} \in \mathbb{R}^{\sqrt{d} \times \sqrt{d}}$, which for the base model is the output of semantic composition.

\begin{exe}
	\ex
	$H_{inner} = tanh(W_{\textsc{comb}} H_{l} + B_{\textsc{comb1}})$
	\label{ex:composition1}
	\ex
	$H_{cand} = tanh(H_{inner} H_{r} + B_{\textsc{comb2}})$
	\label{ex:composition2}
\end{exe}

As in CMS, the primary mode of semantic composition is matrix multiplication. However, LMS improves on CMS in that it avoids associativity. LMS differs from MV-RNN in that it does not learn a $d \times d$ matrix for each vocabulary item. Compared to RNTN, LMS transmits a larger number of activations across layers, given the same parameter count. In both models, the size of the third-order tensor is the dominant factor in determining the number of model parameters. The parameter count of LMS is approximately proportional to the number of activations ($d$), but the parameter count of RNTN is approximately proportional to the cube of the number of activations ($d^{3}$). Therefore, LMS can transmit the same number of activations with fewer model parameters.

\subsection{LMS augmented with LSTM components}
\label{subsec:full}

We augment the base model with LSTM components (LMS-LSTM) to circumvent the problem of long-term dependencies. As in the case of TreeLSTM, we additionally manage cell states ($\vec{c}_{l}$, $\vec{c}_{r}$). Since the LSTM components operate on vectors, we reshape $H_{cand}$, $H_{l}$, and $H_{r}$ into $d \times 1$ column vectors respectively, and produce $\vec{g}$, $\vec{h}_{l}$, and $\vec{h}_{r}$. The output of the LSTM components are calculated based on these vectors, and is reshaped back to a $\sqrt{d} \times \sqrt{d}$ matrix (eq. \ref{ex:lstm reshape}).

\begin{exe}
	\ex
	$\vec{g} = \textsc{vectorize}(H_{cand})$
	\label{ex:candidate}
	\ex
	$\vec{h}_{l} = \textsc{vectorize}(H_{l})$
	\ex
	$\vec{h}_{r} = \textsc{vectorize}(H_{r})$
	\label{ex:reshape}
	\ex
	\hspace{-1pt}$\left[ \begin{array}{c} \vec{i}\\ \vec{f}_{l}\\ \vec{f}_{r}\\ \vec{o} \end{array} \right]
	= \left[ \begin{array}{c} \sigma\\ \sigma\\ \sigma\\ \sigma \end{array} \right]
	\left( W \left[ \begin{array}{c} \vec{h}_{l}\\ \vec{h}_{r} \end{array} \right] + B \right)$
	\label{ex:gates}
	\ex
	$\vec{c} = \vec{f}_{l} \odot \vec{c}_{l} + \vec{f}_{r} \odot \vec{c}_{r} + \vec{i} \odot \vec{g}$
	\label{ex:cell state}
	\ex
	$\vec{h} = \vec{o} \odot tanh(\vec{c})$
	\label{ex:lstm output}
	\ex
	$H = \textsc{to-matrix}(\vec{h})$
	\label{ex:lstm reshape}
\end{exe}

\subsection{Simplified variants}
\label{subsec:simplified}

We implement two LMS-LSTM variants with a simpler composition function as an ablation study. The first variant replaces the equations in (\ref{ex:composition1}) and (\ref{ex:composition2}) with a single equation (eq. \ref{ex:simple1}), which does not utilize a weight matrix. It simply multiplies the matrix representations of two child nodes $H_{l}, H_{r} \in \mathbb{R}^{\sqrt{d} \times \sqrt{d}}$, adds a bias $B_{\textsc{comb}} \in \mathbb{R}^{\sqrt{d} \times \sqrt{d}}$, and feeds the result into a non-linear activation function.

\begin{exe}
	\ex
	$H_{cand} = tanh(H_{l} H_{r} + B_{\textsc{comb}})$
	\label{ex:simple1}
\end{exe}

The second variant is more complex than the first, in a way that a weight matrix $W_{\textsc{comb}} \in \mathbb{R}^{\sqrt{d} \times \sqrt{d}}$ is added to the equation (eq. \ref{ex:simple2}). But unlike the full LMS-LSTM which has two \textit{tanh} layers, it only utilizes one.

\begin{exe}
	\ex
	$H_{cand} = tanh(W_{\textsc{comb}}H_{l}H_{r} + B_{\textsc{comb}})$
	\label{ex:simple2}
\end{exe}

\section{Experiments}
\label{sec:experiments}

\subsection{Implementation details}

As our interest is in the performance of composition functions, we compare LMS-LSTM with TreeLSTM, the previous best known composition function for tree-structured models. To allow for efficient batching, we use the SPINN-PI-NT approach \cite{Bowman2016a}, which implements standard TreeLSTM using stack and buffer data structures borrowed from parsing, rather than tree structures. We implement our model by replacing SPINN-PI-NT's composition function with ours and adding the \textsc{lift} layer.

We use the 300D reference GloVe vectors \citep[840B token version;][]{Pennington2014} for word representations. We fine-tune the word embeddings for improved results. We follow \citet{Bowman2016a} and other prior work in our use of an MLP with product and difference features to classify pairs of sentences.

\begin{exe}
	\ex
	$\vec{x}_{classifier} = \left[ \begin{array}{c} \vec{h}_{premise}\\ \vec{h}_{hypothesis}\\ \vec{h}_{premise} - \vec{h}_{hypothesis}\\ \vec{h}_{premise} \odot \vec{h}_{hypothesis} \end{array} \right]$
	\label{classifier}
\end{exe}

\begin{table*}[t]
	\small
	\begin{center}
		\begin{tabular}{ lrrrrrr }
			\toprule
			\bf Model & \bf Params. & \bf S tr. & \bf S te. & \bf M tr. & \bf M te. mat. & \bf M te. mism.\\
			\midrule
			\multicolumn{7}{c}{\bf Baselines}\\
			\midrule
			CBOW \cite{Williams2017} & -- & -- & 80.6 & -- & 65.2 & 64.6\\
			BiLSTM \cite{Williams2017} & 2.8m & -- & 81.5 & -- & 67.5 & 67.1\\
			Shortcut-Stacked BiLSTM \cite{Nie2017} & 34.7m & -- & 86.1 & -- & 74.6 & 73.6\\
			DIIN \cite{Gong2018} & -- & -- & \textbf{88.0} & -- & \textbf{78.8} & \textbf{77.8}\\
			\midrule
			\multicolumn{7}{c}{\bf Existing Tree-Structured Model Runs}\\
			\midrule
			300D TreeLSTM \cite{Bowman2016a} & 3.4m & 84.4 & 80.9 & -- & -- & --\\
			300D SPINN-PI \cite{Bowman2016a} & 3.7m & 89.2 & 83.2 & -- & -- & --\\
			\midrule
			\multicolumn{7}{c}{\bf Our Experiments}\\
			\midrule
			441D LMS (base) & 2.0(+11.6m) & 79.7 & 76.5 & -- & -- & --\\
			441D LMS-LSTM (simplified, $- W_{\textsc{comb}}, - tanh$) & 3.3(+11.6)m & 90.5 & 84.1 & -- & -- & --\\
			324D LMS-LSTM (simplified, $- tanh$) & 2.2(+11.6)m & 92.5 & 84.5 & -- & -- & --\\
			\midrule
			700D TreeLSTM & 2.0(+11.6)m & 89.5 & 83.6 & -- & -- & --\\
			576D LMS-LSTM (full) & 4.6(+11.6)m & 86.0 & \underline{84.9} & -- & -- & --ß\\
			700D TreeLSTM & 4.6(+30.2)m & -- & -- & 78.9 & 70.0 & 69.7\\
			576D LMS-LSTM (full) & 5.9(+30.2)m & -- & -- & 80.5 & \underline{71.3} & \underline{71.6}\\
			\bottomrule
		\end{tabular}
		\caption{Results on NLI classification with sentence-to-vector encoders. \textit{Params.} is the approximate number of model parameters, and the numbers in parentheses indicate the parameters contributed by word embeddings. \textit{S tr.}, and \textit{S te.} are the class accuracies (\%) on SNLI train set and test set, respectively. \textit{M tr.}, \textit{M te. mat.}, and \textit{M te. mism.} are the class accuracies (\%) on MultiNLI train set, matched test set, and mismatched test set, respectively. Underlining marks the best result among tree-structured models.}
		\label{tab:experiment snli}
	\end{center}
\end{table*}

The feature vector is fed into an MLP that consists of two ReLU neural network layers and a softmax layer. In both models, the objective function is a sum of a cross-entropy loss function and an L2 regularization term. Both models use the Adam optimizer \cite{Kingma2014}. Dropout \cite{Srivastava2014} is applied to the classifier and to the word embeddings. The MLP layer also utilizes Layer Normalization \cite{Ba2016}.\footnote{The source code and the checkpoints for the models trained for the NLI tasks are available at \url{https://github.com/nyu-mll/spinn}.}

\subsection{Datasets}

We first train and test our models on the \textit{Stanford Natural Language Inference corpus} \citep[SNLI;][]{Bowman2015}. The SNLI corpus contains 570,152 pairs of natural language sentences that are labeled for \textit{entailment}, \textit{contradiction}, and \textit{neutral}. It consists of sentences that were written and validated by humans. Along with the MultiNLI corpus introduced below, it is two orders of magnitude larger than other human-authored resources for NLI tasks. The following example illustrates the general format of the corpus.

\begin{exe}
	\item \hspace{-1em} 
    \textsc{\textbf{premise:}} A soccer game with multiple males playing.\\
	\textsc{\textbf{hypothesis:}} Some men are playing a sport.\\
	\textsc{\textbf{label:}} Entailment
	\label{ex:snli}
\end{exe}

We test our models on the \textit{Multi-Genre Natural Language Inference corpus} \citep[MultiNLI;][]{Williams2017}. The corpus consists of 433k pairs of examples, and each pair is labeled for \textit{entailment}, \textit{contradiction}, and \textit{neutral}. MultiNLI has the same format as SNLI, so it is possible to train on both datasets at the same time (as we do when testing on MultiNLI). Two notable features distinguish MultiNLI from SNLI: (i) It is collected from ten distinct genres of spoken and written English. This makes the dataset more representative of human language use. (ii) The examples in MultiNLI are considerably longer than the ones in SNLI. These two features make MultiNLI classification fairly more difficult than SNLI. The pair of sentences in (\ref{ex:multinli}) is an illustrative example. The sentences are from the section of the corpus that is transcribed verbatim from telephone speech. 

\begin{exe}
	\item \hspace{-0.8em} \textsc{\textbf{genre:}} Telephone speech\\
	\textsc{\textbf{premise:}} Yes now you know if if everybody like in August when everybody's on vacation or something we can dress a little more casual or\\
	\textsc{\textbf{hypothesis:}} August is a black out month for vacations in the company.\\
	\textsc{\textbf{label:}} Contradiction
	\label{ex:multinli}
\end{exe}

The MultiNLI training set consists of five different genres of spoken and written English, the \textit{matched} test set contains sentence pairs from only those five genres, and the \textit{mismatched} test set contains sentence pairs from additional genres.

We also experiment on the Stanford Sentiment Treebank \citep[SST;][]{Socher2013}, which is constructed by extracting movie review excerpts written in English from \url{rottentomatoes.com}, and labeling them with Amazon Mechanical Turk. Each example in SST is paired with a parse tree, and each node of the tree is tagged with a fine-grained sentiment label (5 classes).

\section{Results and Analysis} 

Table~\ref{tab:experiment snli} summarizes the results on SNLI and MultiNLI classification. We use the same preprocessing steps for all results we report, including loading the parse trees supplied with the datasets. Dropout rate, size of activations, number and size of MLP layers, and L2 regularization term are tuned using repeated random search. MV-RNN and RNTN introduced in the earlier sections are extremely expensive in terms of computational resources, and training the models with comparative hyperparameter settings quickly runs out of memory on a high-end GPU. We do not include them in the comparison for this reason. TreeLSTM performs the best with one MLP layer, while LMS-LSTM displays the best performance with two MLP layers. The difference in parameter count is largely affected by this choice, and in principle one model does not demand notably more computational resources than the other.

On the SNLI test set, LMS-LSTM has an additional 1.3\% gain over TreeLSTM. Also, both of the simplified variants of LMS-LSTM outperform TreeLSTM, but by a smaller margin. On the MultiNLI test sets, LMS-LSTM scores 1.3\% higher on the matched test set and 1.9\% higher on the mismatched test set.

We cite the state-of-the-art results of non-tree-structured models, although these models are only relevant for our absolute performance numbers. The Shortcut-Stacked sentence encoder achieves the state-of-the-art result among non-attention-based models, outperforming LMS-LSTM. While this paper focuses on the design of the composition function, we expect that adding depth along the lines of \citet{Irsoy2014} and shortcut connections to LMS-LSTM would offer comparable results. \citepos{Gong2018} attention-based Densely Interactive Inference Network (DIIN) displays the state-of-the-art performance among all models. Applying various attention mechanisms to tree-structured models is left for future research.

\begin{table}[t]
	\small
	\begin{center}
		\begin{tabular}{ lrrrr }
			\toprule
			& \multicolumn{2}{c}{LMS-LSTM} & \multicolumn{2}{c}{TreeLSTM} \\ 
			\textbf{Phenomenon} & \textbf{Mat.} & \textbf{Mismat.} & \textbf{Mat.} & \textbf{Mismat.} \\
			\midrule
			Pronoun & \textbf{72.0} & \textbf{71.6} & 69.6 & 70.3 \\
			Quantifier & \textbf{72.2} & \textbf{71.7} & 69.9 & 70.5 \\
			Modal & \textbf{70.6} & \textbf{70.8} & 69.8 & 69.2 \\
			Negation & \textbf{72.3} & \textbf{74.1} & 70.3 & 72.4 \\
			Wh-term & \textbf{70.5} & \textbf{69.7} & 68.6 & 68.6 \\
			Belief verb & \textbf{70.1} & \textbf{70.1} & 68.4 & 68.7 \\
			Time term & \textbf{70.0} & \textbf{71.1} & 67.3 & 69.4 \\
			Discourse mark. & \textbf{68.8} & \textbf{68.8} & 67.0 & 67.0 \\
			Presup. triggers & \textbf{71.5} & \textbf{71.9} & 69.1 & 69.9 \\
			Compr./Supr. & \textbf{69.0} & \textbf{67.5} & 67.0 & 67.1 \\
			Conditionals & \textbf{69.7} & \textbf{71.3} & 68.2 & 70.5 \\
			Tense match & \textbf{73.3} & \textbf{72.5} & 71.0 & 71.2 \\
			Interjection & \textbf{69.7} & \textbf{74.3} & \textbf{69.7} & 72.5 \\
			Adj/Adv & \textbf{72.6} & \textbf{72.0} & 70.3 & 70.7 \\
			Determiner & \textbf{72.4} & \textbf{72.1} & 70.3 & 70.8 \\
			Length 0-10 & \textbf{72.8} & \textbf{72.8} & 69.8 & 72.7 \\
			Length 11-14 & \textbf{72.6} & \textbf{72.8} & 72 & 70.5 \\
			Length 15-19 & \textbf{71.0} & \textbf{70.8} & 69.3 & 68.3 \\
			Length 20+ & \textbf{75.2} & 68.2 & 69.8 & \textbf{69.0} \\
			\bottomrule
		\end{tabular}
		\caption{MultiNLI development set classification accuracies (\%), classified using the tags introduced in \citet{Williams2017}.}
		\label{tab:experiment linguistic phenomenon}
	\end{center}
\end{table}

We inspect the performance of the models on certain subsets of the MultiNLI corpus that manifest linguistically difficult phenomena, which was categorized by \citet{Williams2017}. The phenomena include pronouns, quantifiers (e.g., \textit{every}, \textit{each}, \textit{some}), modals (e.g., \textit{must}, \textit{can}), negation, wh-terms (e.g., \textit{who}, \textit{where}), belief verbs (e.g., \textit{believe}, \textit{think}), time terms (e.g., \textit{then}, \textit{today}), discourse markers (e.g., \textit{but}, \textit{thus}), presupposition triggers (e.g., \textit{again}, \textit{too}), and so on. In linguistic semantics, these phenomena are known to involve complex interactions that are more intricate than a simple merger of concepts. For instance, modals express possibilities or necessities that are beyond ``here and now''. One can say `John might be home' to convey that there is a possibility that John is home. The utterance is perfectly compatible with a situation in which John is in fact at school, so modals like \textit{might} let us reason about things that are potentially false in the real world. We use the same code as \citet{Williams2017} to categorize the data to make fair comparisons.

In addition to the categories offered by \citet{Williams2017}, we inspect whether sentences containing adjectives/adverbs affect the performance of the models. \citet{Baroni2010} show that adjectives are better represented as matrices, as opposed to vectors. We also inspect whether the presence of a determiner in the hypothesis that refers back to a salient referent in the premise affects the model performance. Determiners are known to encode intricate properties in linguistic semantics and have been one of the major research topics \cite{Elbourne2005, Charlow2014}. Lastly, we examine whether the performance of the models varies with respect to sentence length, as longer sentences are harder to comprehend.

Table \ref{tab:experiment linguistic phenomenon} summarizes the result of the inspection on linguistically difficult phenomena. We see gains uniformly across the board, but with particularly clear gains on negation (+2\% on the matched set/+1.7\% on the mismatched set), quantifiers (+2.3\%/+1.2\%), time terms (+2.7\%/+1.7\%), tense matches (+2.3\%/+1.3\%), adjectives/adverbs (+2.3\%/+1.3\%), and longer sentences (length 15-19: +1.7\%/+2.5\%; length 20+: +5.4\%/-0.8\%).

\begin{table}[t]
	\small
	\begin{center}
		\begin{tabular}{ lr }
			\toprule
			\textbf{Model} & \textbf{Test} \\
			\midrule
			\multicolumn{2}{c}{\bf Baselines}\\
			\midrule
			MV-RNN \cite{Socher2013} & 44.4\\
			RNTN \cite{Socher2013} & 45.7\\
			Deep RNN \cite{Irsoy2014} & 49.8\\
			TreeLSTM \cite{Tai2015} & 51.0\\
			TreeBiGRU w/ attention & 52.4\\
			\cite{Kokkinos2017}\\
			\midrule
			\multicolumn{2}{c}{\bf Our Experiments}\\
			\midrule
			312D TreeLSTM & 48.9\\
			144D LMS-LSTM & 50.1\\
			\bottomrule
		\end{tabular}
		\caption{Five-way test set classification accuracies (\%) on the Stanford Sentiment Treebank.}
		\label{tab:experiment sst}
	\end{center}
\end{table}

Table \ref{tab:experiment sst} summarizes the results on SST classification, particularly on the fine-grained task with 5 classes. While our implementation does not exactly reproduce \citepos{Tai2015} TreeLSTM results, a comparison between our trained TreeLSTM and LMS-LSTM is consistent with the patterns seen in NLI tasks.

We examine how well the constituent representations produced by LMS-LSTM and TreeLSTM encode syntactic category information. As mentioned earlier, there is a consensus in linguistic semantics that semantic composition involves function application (i.e., feeding an argument to a function) which goes beyond a simple merger of two concepts. Given that the syntactic category of a node determines whether the node serves as a function or an argument in semantic composition, we hypothesize that the distributed representation of each node would encode syntactic category information if the models learned how to do function application. To assess the quality of the representations, we first split the SNLI development set into training and test sets. From the training set, we extract the hidden state of every constituent (i.e., phrase) produced by the best performing models. For each of the models, we train linear classifiers that learn to do the following two tasks: (i) 3-way classification, which trains and tests exclusively on noun phrases, verb phrases, and prepositional phrases, and (ii) 19-way classification, which trains and tests on all 19 category labels attested in the SNLI development set. The distribution of the 19 category labels is provided in Table \ref{tab:type distribution}. We opt for a linear classifier to keep the classification process simple, so that we can properly assess the quality of the constituent representations.

\begin{table}[t]
	\small
	\begin{center}
		\begin{tabular}{ lrr }
			\toprule
			\textbf{Category} & \textbf{\# of samples} & \textbf{Ratio (\%)} \\
			\midrule
			NP & 63346 & 43.31\\
			VP & 30534 & 20.08\\
			PP & 25624 & 17.98\\
			ROOT & 18267 & 12.49\\
			S & 4863 & 3.06\\
			SBAR & 2004 & 1.20\\
			ADVP & 1136 & 0.27\\
			ADJP & 408 & 0.13\\
			Etc. & 160 & 1.45\\
			\bottomrule
		\end{tabular}
		\caption{Syntactic category distribution of SNLI development set, classified using the tags introduced in \citet{Bowman2015}.}
		\label{tab:type distribution}
	\end{center}
\end{table}

Table \ref{tab:experiment type} summarizes the results on the syntactic category classification task. As a baseline, we train a bag-of-words (BOW) model which produces the hidden state of a given phrase by summing the GloVe embeddings of the words of the phrase. We train and test on the hidden states produced by BOW as well. The hidden state representations produced by LMS-LSTM yield the best results on both 3-way and 19-way classification tasks. Comparing LMS-LSTM and TreeLSTM representations, we see a 5.1\% gain on the 3-way classification and a 5.5\% gain on the 19-way classification.

\begin{table}[t]
	\small
	\begin{center}
		\begin{tabular}{ lrrrr }
			\toprule
			& \multicolumn{2}{c}{\textbf{3-way}} & \multicolumn{2}{c}{\textbf{19-way}} \\
			\textbf{Model} & \textbf{Train} & \textbf{Test} & \textbf{Train} & \textbf{Test} \\
			\midrule
			300D BOW & 86.4 & 85.6 & 82.7 & 82.1 \\
			700D TreeLSTM & 93.2 & 91.2 & 90.0 & 86.6 \\
			576D LMS-LSTM & 97.3 & \textbf{96.3} & 94.0 & \textbf{92.1} \\
			\bottomrule
		\end{tabular}
		\caption{Syntactic category classification accuracies (\%) on SNLI development set, classified using the tags introduced in \citet{Bowman2015}.}
		\label{tab:experiment type}
	\end{center}
\end{table}

\begin{figure}[t]
	\begin{center}
		\includegraphics[width=7.5cm]{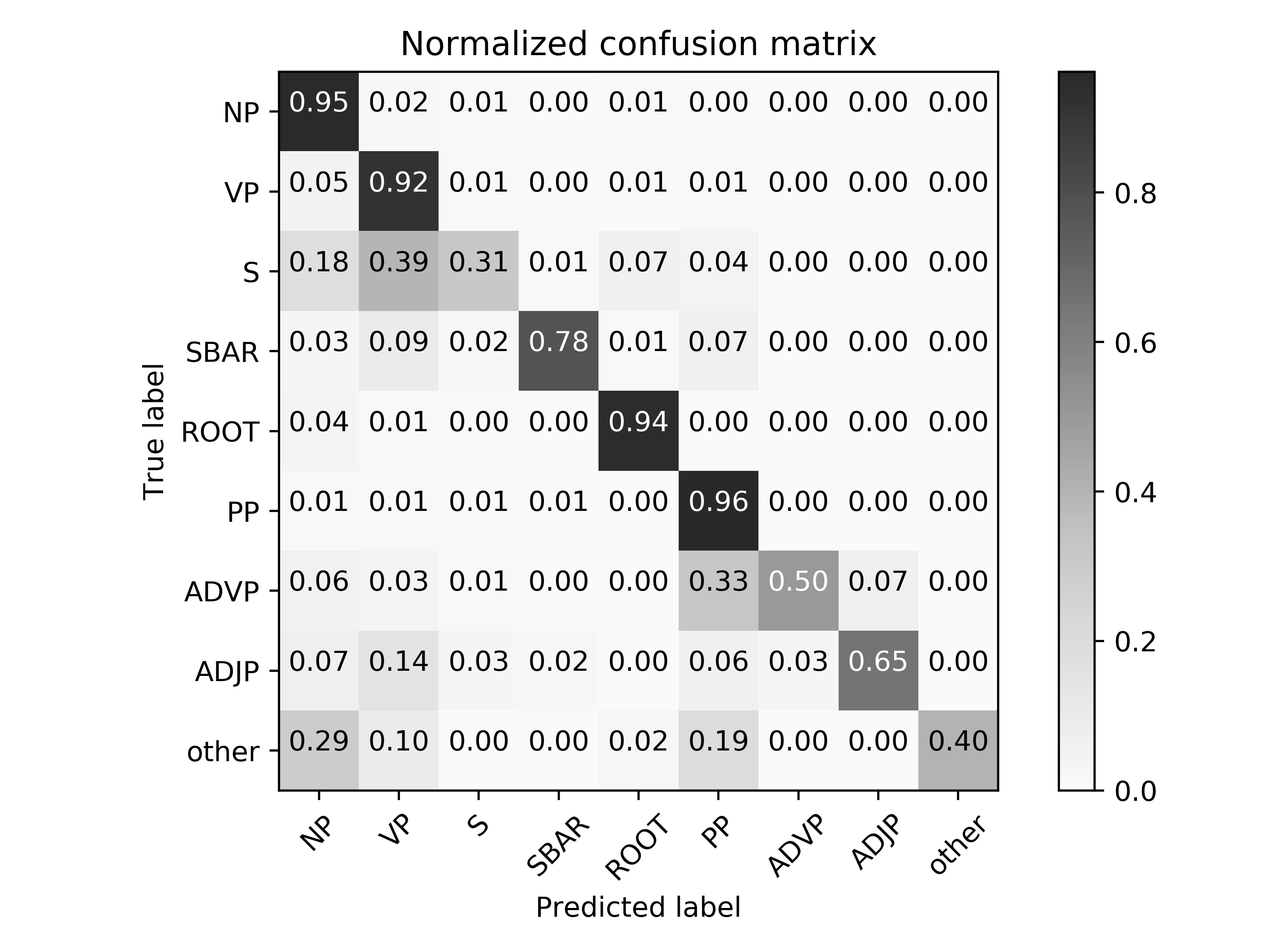}\\
		{\small (a) LMS-LSTM\\}
		\includegraphics[width=7.5cm]{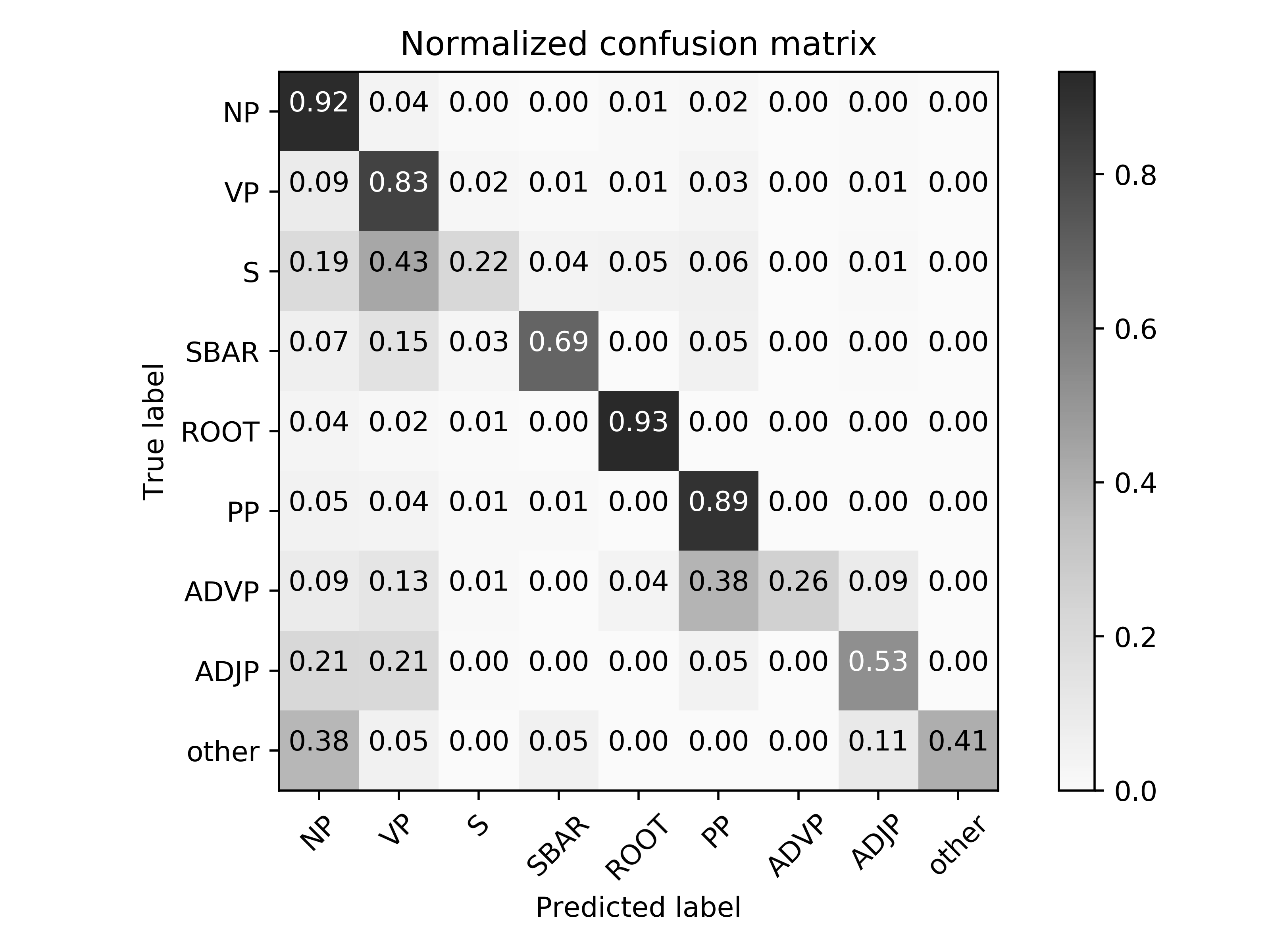}\\
		{\small (b) TreeLSTM\\}
		\caption{Confusion matrices of LMS-LSTM and TreeLSTM for 19-way linear classification.}
		\label{fig:category}
	\end{center}
\end{figure}

Figure \ref{fig:category} depicts the corresponding confusion matrices for the 19-way classification task. We show the most frequent eight classes due to space limitations. We observe notable gains on adverbial phrases (ADVP; +24\%), adjectival phrases (ADJP; +12\%), verb phrases (VP; +9\%), and clauses introduced by subordinate conjunction (SBAR; +9\%). We also observe a considerable gain on non-terminal declarative clauses (S; +9\%), although the absolute number is fairly low compared to other categories. While we do not have a full comprehension of the drop in classification accuracy, we speculate that the ambiguity of infinitival clauses and gerund phrases is one of the culprits. As exemplified in (\ref{ex:dog}) and (\ref{ex:women}) respectively, infinitival clauses and gerund phrases are not only labeled as a VP but also as an S in the SNLI dataset. Given that our experiment is set up in a way that each constituent is assigned exactly one category label, a good number of infinitival clauses and gerund phrases that are labeled as an S could have been classified as a VP, resulting in a drop in classification accuracy. On the other hand, VP constituents are less affected by the ambiguity because the majority of them are neither an infinitival clause nor a gerund phrase, as shown in (\ref{ex:fish}).

\begin{exe}
	\ex A dog carries a snowball [$_{\textsc{S}}$ [$_{\textsc{VP}}$ to give it to his owner ]]
	\label{ex:dog}
	
	\ex Two women are embracing while [$_{\textsc{S}}$ [$_{\textsc{VP}}$ holding to-go packages after just eating lunch ]]
	\label{ex:women}

	\ex A few people [$_{\textsc{VP}}$ are [$_{\textsc{VP}}$ catching fish ]]
	\label{ex:fish}
\end{exe}

With the exclusion of non-terminal declarative clauses, the categories we see notable gains are known to play the role of a function in semantic composition. On the other hand, both models are efficient in identifying noun phrases (NP), which are typically arguments of a function in semantic composition. We speculate that the results are indicative of LMS-LSTM's ability to identify functions and arguments, and this hints that the model is learning to do function application.



\section{Conclusion}
\label{sec:conclusion}

In this paper, we propose a novel model for semantic composition that utilizes matrix multiplication. Experimental results indicate that, while our model does not reach the state of the art on any of the three datasets under study, it does substantially outperform all known tree-structured models, and lays a strong foundation for future work on tree-structured compositionality in artificial neural networks. 

\section*{Acknowledgments}

We thank NVIDIA Corporation for the donation of the Titan X Pascal GPU used for this research. This project has benefited from financial support to SB by Google, Tencent Holdings, and Samsung Research.

\bibliographystyle{acl_natbib_nourl}
\bibliography{mendeley}

\end{document}